\documentclass[letterpaper, 10 pt, conference]{ieeeconf}

\usepackage{xcolor}
\definecolor{maincolor}{HTML}{5b9bd5}
\usepackage[colorlinks=true,
            allcolors=maincolor]{hyperref}
\usepackage[noadjust,sort]{cite}
\usepackage{mathtools}
\usepackage{amsmath}
\usepackage{amssymb}
\usepackage[font=footnotesize]{caption}
\usepackage{subcaption}
\usepackage{url}
\usepackage{etoolbox}
\usepackage{booktabs}
\usepackage{balance}

\IEEEoverridecommandlockouts
\newcommand{\name}{\textit{EquivAct}}

\title{\LARGE \bf
EquivAct: SIM(3)-Equivariant Visuomotor Policies\\ beyond Rigid Object Manipulation
}

\author{Jingyun Yang$^{*1}$, Congyue Deng$^{*1}$, Jimmy Wu$^{2}$, Rika Antonova$^{1}$, Leonidas Guibas$^{1}$, Jeannette Bohg$^{1}$ 
\thanks{$^*$ Equal contribution. $^1$ Department of Computer Science, Stanford University. $^2$ Department of Computer Science, Princeton University.}%
\thanks{This work was supported in part by the Toyota Research Institute and the National Science Foundation grant No.2030859 to the Computing Research Association for the CIFellows Project.}
\thanks{\scriptsize{Contact: \texttt{\{jingyuny, rika.antonova\}@stanford.edu}}}%
}

\usepackage{color}
\usepackage{xcolor}
\definecolor{turquoise}{cmyk}{0.65,0,0.1,0.3}
\definecolor{purple}{rgb}{0.65,0,0.65}
\definecolor{dark_green}{rgb}{0, 0.5, 0}
\definecolor{orange}{rgb}{0.8, 0.6, 0.2}
\definecolor{red}{rgb}{0.8, 0.2, 0.2}
\definecolor{darkred}{rgb}{0.6, 0.1, 0.05}
\definecolor{blueish}{rgb}{0.0, 0.3, .6}
\definecolor{light_gray}{rgb}{0.7, 0.7, .7}
\definecolor{pink}{rgb}{1, 0, 1}
\definecolor{greyblue}{rgb}{0.25, 0.25, 1}
\definecolor{tab_blue}{HTML}{1f77b4}
\definecolor{tab_orange}{HTML}{ff7f0e}
\definecolor{LightRed}{rgb}{0.99,0.89,0.89}

\definecolor{mesh_misty_rose}{HTML}{e6aaa3}
\definecolor{mesh_yellow}{HTML}{ffba00}

\usepackage{amsmath,amsfonts,bm}

\def\eqref#1{equation~\ref{#1}}

\def\1{\bm{1}}

\def\rvm{{\mathbf{m}}}

\def\rvt{{\mathbf{t}}}

\def\rvv{{\mathbf{v}}}

\def\rvx{{\mathbf{x}}}
\def\rvy{{\mathbf{y}}}

\def\rmM{{\mathbf{M}}}

\def\rmR{{\mathbf{R}}}

\def\rmT{{\mathbf{T}}}

\def\rmV{{\mathbf{V}}}

\def\rmX{{\mathbf{X}}}

\DeclareMathAlphabet{\mathsfit}{\encodingdefault}{\sfdefault}{m}{sl}
\SetMathAlphabet{\mathsfit}{bold}{\encodingdefault}{\sfdefault}{bx}{n}

\def\gL{{\mathcal{L}}}

\def\sR{{\mathbb{R}}}

\newcommand{\SIMthree}{\mathrm{SIM}(3)}

\newcommand{\tinv}{\text{inv}}

\begin{document}

\newcommand{\insertteaser}{
    \vspace{5px}
    \includegraphics[width=0.8\linewidth]{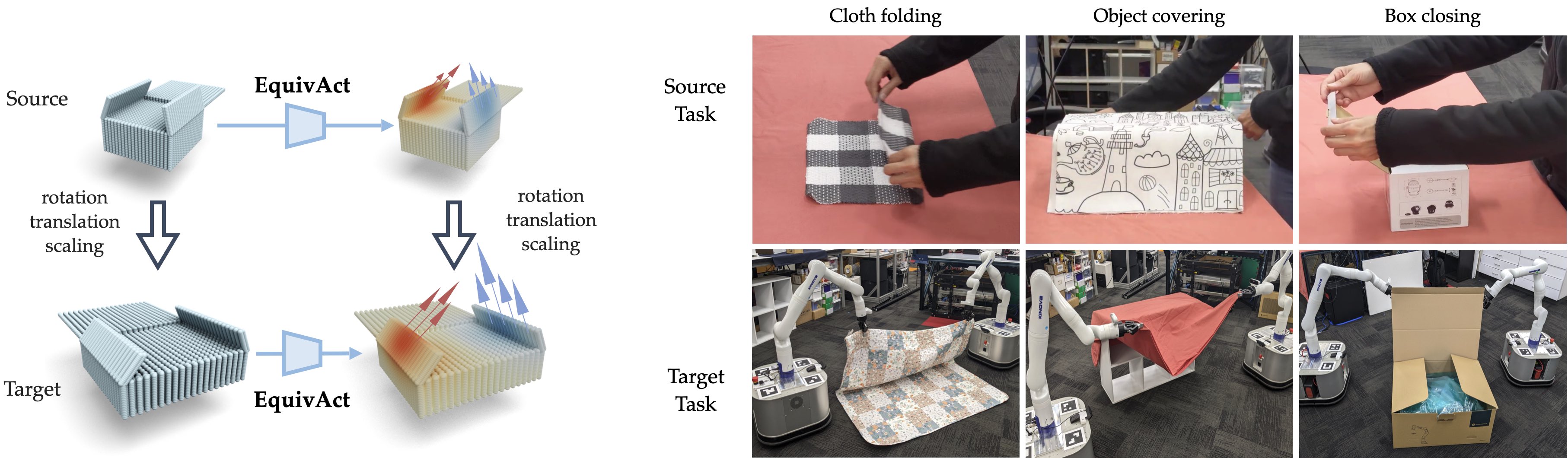}
    \captionof{figure}{\textbf{Overview.} Constructed with $\SIMthree$-equivariant point cloud networks, our method takes a few examples of solving a source task, then generalizes zero-shot to changes in object \textit{appearances}, \textit{scales}, and \textit{poses}.}
    \label{fig:teaser}
    \vspace{-5px}
}
\let\oldcite\cite
\renewcommand*\cite[1]{{\color{maincolor}\oldcite{#1}}}

\makeatletter
\apptocmd{\@maketitle}{\centering\insertteaser}{}{}
\makeatother

\maketitle
\setcounter{figure}{1}
\thispagestyle{empty}
\pagestyle{empty}

\begin{abstract}
If a robot masters folding a kitchen towel, we would expect it to master folding a large beach towel. However, existing policy learning methods that rely on data augmentation still don't guarantee such generalization. Our insight is to add equivariance to both the visual object representation and policy architecture. We propose \textit{EquivAct} which utilizes SIM(3)-equivariant network structures that guarantee generalization across all possible object translations, 3D rotations, and scales by construction. EquivAct is trained in two phases. We first pre-train a SIM(3)-equivariant visual representation on simulated scene point clouds. Then, we learn a SIM(3)-equivariant visuomotor policy using a small amount of source task demonstrations. We show that the learned policy directly transfers to objects that substantially differ from demonstrations in scale, position, and orientation. We evaluate our method in three manipulation tasks involving deformable and articulated objects, going beyond typical rigid object manipulation tasks considered in prior work.  We conduct experiments both in simulation and in reality. For real robot experiments, our method uses 20 human demonstrations of a tabletop task and transfers zero-shot to a mobile manipulation task in a much larger setup. Experiments confirm that our contrastive pre-training procedure and equivariant architecture offer significant improvements over prior work. 
Project website: \href{https://equivact.github.io}{\texttt{equivact.github.io}}
\end{abstract}
\section{Introduction}

Given a few examples of how to solve a manipulation task, humans can extrapolate and learn to solve variations of the same task where objects have different visual or physical properties. Existing works in robot learning still require extensive data augmentation to make the learned policies generalize to varied object scales, orientations, and visual appearances \cite{james2019sim, mehta2020active, laskin2020reinforcement, hansen2021generalization, yu2023scaling}. Even then, augmentations do not guarantee generalization to unseen variations.

In this work, we focus on the problem of learning visuomotor policies that can take a few example trajectories from a single source manipulation scenario as input, then generalize zero-shot to scenarios with changes in objects' appearances, scales, and poses. We go beyond pick-and-place of rigid objects and also handle deformable and articulated objects, such as clothes and boxes. Our insight is to \emph{add equivariance to both the visual object representation and policy architecture}, enabling policies to generalize to novel object positions, orientations, and scales \emph{by construction}. 

We propose \name, a novel visuomotor policy learning method that can learn closed-loop policies for 3D robot manipulation tasks using demonstrations from a single source manipulation scenario, then generalize zero-shot to new variations. A resulting policy takes a partial point cloud of the scene and robot end-effector poses as input, then outputs robot actions, which include end-effector velocity and gripper commands. Different from neural network architectures used in most prior works, we employ SIM(3)-equivariant network structures. This implies that when the input point cloud and end-effector positions are translated, rotated and scaled, the output end-effector velocities transform accordingly. With this, our method can learn from demonstrations of small-scale tabletop tasks, then offer zero-shot generalization to mobile manipulation tasks that involve larger variants of objects with different appearances. 

Our method is composed of two phases: a representation learning phase and a policy learning phase. 
The representation learning phase serves two purposes. First, although the proposed architecture is equivariant to uniform scaling, we need to pre-train the representation to handle non-uniform scaling. Second, although rotation is handled via equivariance, we need robustness to nonlinear changes caused by objects' self-occlusions when altering camera view angles.
Hence, for the representation learning phase we collect a set of simulated point clouds of objects with randomized non-uniform re-scaling. This data is not a target task demo and does not include robot actions. We train a SIM(3)-equivariant encoder-decoder architecture~\cite{lei2023efem} that takes a single-view partial scene point cloud as input, then outputs both global and local features.
We use a contrastive learning loss on paired point clouds, which ensures that local features for corresponding object parts in similar poses are closer than disparate parts.
Next, in the policy learning phase, we assume access to a small set of demonstrations of a human solving a source task. Using these, we train a closed-loop policy that takes a partial point cloud of the scene as input, extracts global and local features using the encoder from the representation learning phase, and then passes the features through a SIM(3)-equivariant action prediction network to obtain end-effector actions.

We evaluate our method in three challenging scenarios that go beyond typical rigid object manipulation tasks of prior work~\cite{simeonov2022neural,simeonov2023se,xue2023useek,ryu2022equivariant,weng2023neural}:  comforter folding, container covering, and box closing (see Fig. \ref{fig:teaser}).
For each scenario, we show that our approach learns a closed-loop policy from source demonstrations, and then completes the target task without any further adaptation. We also compare with methods that rely on extensive augmentations and demonstrate that our approach of leveraging equivariance is more efficient for ensuring generalization to out-of-distribution object poses and scales.
\section{Related Work}

\subsection{Equivariant 3D Learning}

A number of 3D deep learning works have studied the construction of neural network architectures that are equivariant to 3D transformations such as rigid transformations \cite{thomas2018tensor, fuchs2020se, chen2021equivariant, deng2021vector, assaad2022vn, katzir2022shape, li2022orientation, poulenard2021functional}, multi-part motions \cite{deng2023banana, yu2022rotationally, lei2023efem, liu2023self}, or more general and arbitrary group actions \cite{kondor2018generalization, cohen2019general, weiler2021coordinate, aronsson2022homogeneous, xu2022unified}.
Such network designs enable natural and robust generalizations to out-of-distribution inputs by construction without additional data augmentation in the training process.
In our work, we leverage the implicit shape representation from \cite{lei2023efem} powered by the network structures from \cite{deng2021vector} to facilitate more geometrically interpretable and generalizable feature learning.

\subsection{Equivariant Representations for Robot Manipulation}

Prior works studied equivariant features for robot manipulation \cite{simeonov2022neural,simeonov2023se,xue2023useek,ryu2022equivariant,weng2023neural}, but had several limitations: (1) focused on pick-and-place tasks and didn't handle harder tasks with articulated and deformable objects; (2) mostly based on open-loop policies or hand-designed pick-and-place primitives, lacking closed-loop feedback; (3) only handled equivariances in translation and rotation and didn't consider scale equivariance. \cite{jia2023seil} combines SO(2)-equivariant closed-loop policy learning with additional simulated augmentations for a suite of tabletop rigid object manipulation tasks, but cannot handle out-of-distribution objects with unseen uniform and non-uniform scaling. Compared to prior works, our work learns a robot manipulation policy that supports closed-loop feedback, can handle deformable and articulated objects, is equivariant to translation, rotation, and uniform scaling by construction, and achieves generalization for objects with non-uniform scaling with a representation learning phase.

\subsection{3D Representations for Deformable and Articulated Object Manipulation}

The problem of 3D manipulation of deformable and articulated objects has been studied by several prior works, including FlingBot \cite{ha2022flingbot}, GarmentNets \cite{chi2021garmentnets}, FabricFlowNet \cite{weng2022fabricflownet}, and ACID \cite{shen2022acid}. However, these works often are very focused on specific manipulation problems. For example, FlingBot focuses only on the task of flinging, FabricFlowNet is designed for pick-and-place tasks with cloth, and ACID is designed to manipulate volumetric deformable objects. In contrast, we propose a framework capable of learning a variety of 3D manipulation tasks, including ones that involve deformable and articulated objects.
\section{Method}

\begin{figure}
    \vspace{5px}
    \centering
    \includegraphics[width=0.45\textwidth]{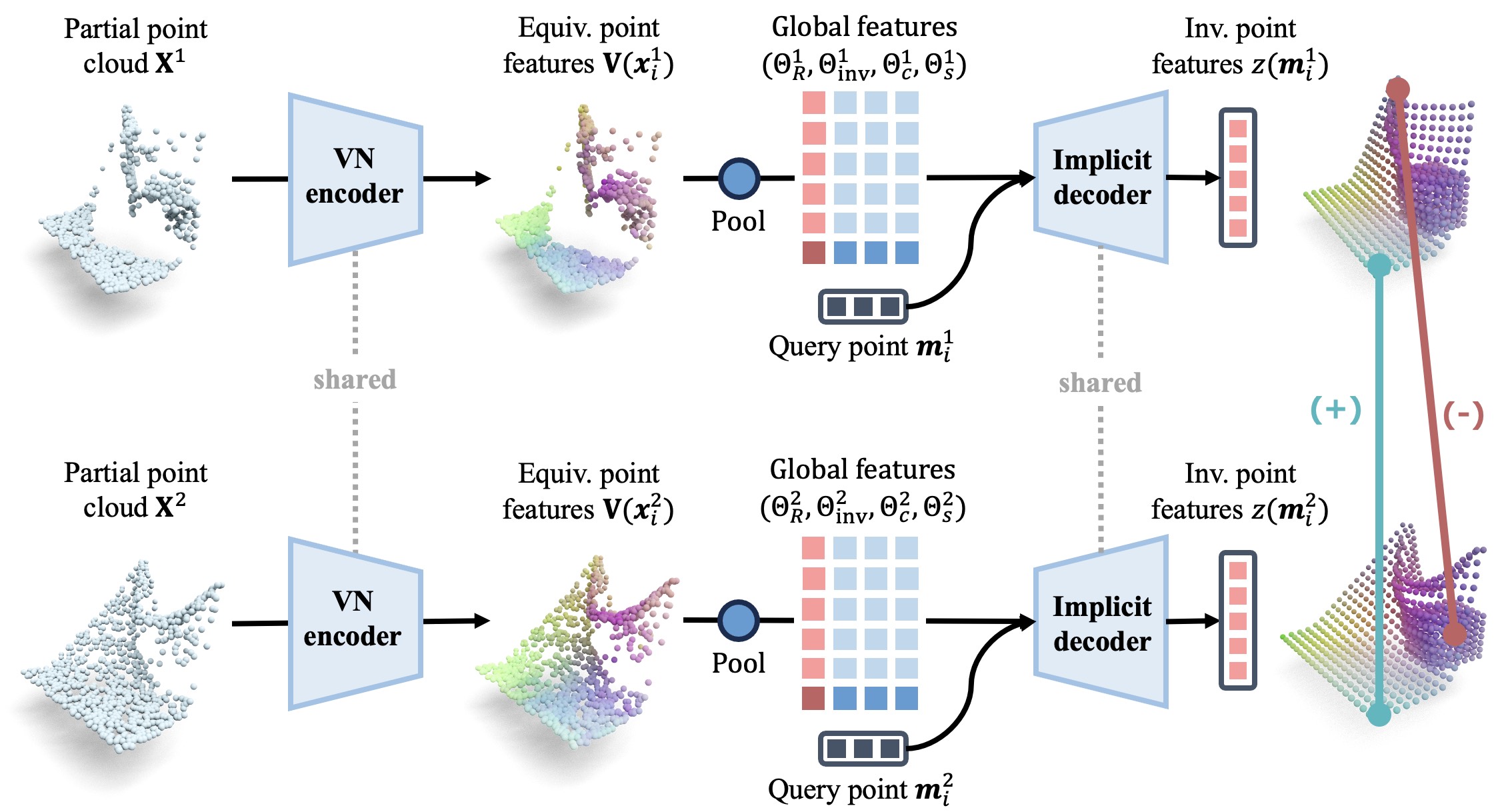}
    \caption{\textbf{Representation learning pipeline.} This phase takes paired partial point clouds as inputs, processes them through an equivariant encoder-decoder architecture, then employs a contrastive loss based on invariant point features, yielding equivariant global and local features as output.}
    \label{fig:rep_learning}
    \vspace{-12pt}
\end{figure}

\subsection{Preliminaries}

\textbf{SIM(3)-equivariance.}
Consider a function $f$ that takes a point cloud $\rmX\!\in\!\sR^{N\times3}$ as input. We say it is $\SIMthree$-equivariant if, for any rigid 3D transformation $\rmT \!:=\! (\rmR,\rvt, s) \in\SIMthree$ with rotation $\rmR$, translation $\rvt$, and scale $s$, the output of $f$ transforms coherently with the input, that is $f(\rmT\rmX) = \rmT f(\rmX)$.
In \cite{deng2021vector}, the authors introduced `Vector Neurons' (VN) for constructing rotation-equivariant point cloud networks.
In \cite{lei2023efem}, VN building blocks were used to obtain a $\SIMthree$-equivariant shape encoding using implicit representations.
Given a partial point cloud $\rmX$, a VN encoder $\Phi$ encodes the input into a latent code $\Theta = \Phi(\rmX)$ comprised of four components $\Theta := (\Theta_R, \Theta_{\text{inv}}, \Theta_c, \Theta_s)$.
Here, $\Theta_R\in\sR^{C\times3}$ is a rotation equivariant latent representation, $\Theta_{\tinv}\in\sR^C$ is an invariant latent representation, scalar $\Theta_s\in\sR$ represents object scale, and $\Theta_c\in\sR^3$ represents the object centroid.
For any $\rmT = (\rmR,\rvt, s) \in \SIMthree$, the equivariance of encoder $\Phi$ can be written as:
\begin{equation}
    \rmT\Theta = (\Theta_R\rmR, \Theta_{\tinv}, s\Theta_c \rmR+\rvt, s\Theta_s) = \Phi(s\rmX\rmR+\rvt).
\end{equation}
The latent code can be decoded by an implicit decoder $\Psi$ that takes a query position $\rvx \in \mathbb{R}^3$ as input and outputs a per-point feature $z(\rvx)$ with:
\begin{equation}
    z(\rvx) = \Psi(\Theta_{\tinv}, \langle \Theta_R, (\rvx - \Theta_c) / \Theta_s \rangle).
\end{equation}

\textbf{Problem formulation.}
We consider a problem where an agent is given a set of demonstrations $\mathcal{D}_\text{demo} = \{\tau_d\}_{d=1}^{N_\text{demo}}$. Each demo $\tau_d$ consists of a sequence of tuples $\{(o_t, a_t)\}_{t=1}^{T}$, where $o_t = (\rmX_t, \{\rvy^{e}_t\}_{e=1}^E)$ is an observation that includes a point cloud of the scene $\rmX_t$ and end-effector poses $\{\rvy^{e}_t\}_{e=1}^E$, while $a_t$ is an action that consists of velocity and gripper commands for all end-effectors.
Using $\mathcal{D}_\text{demo}$ dataset from a source scenario, we train a policy $\pi(a|o)$ to take a point cloud of the target scene as input and output an action $a$. We then deploy the learned policy in a target scenario that involves objects with different appearances, scales, and poses.

\subsection{Learning SIM(3)-equivariant 3D Visual Representations from Simulation Data}

\begin{figure}[t]
    \vspace{5px}
    \centering
    \includegraphics[width=0.485\textwidth]{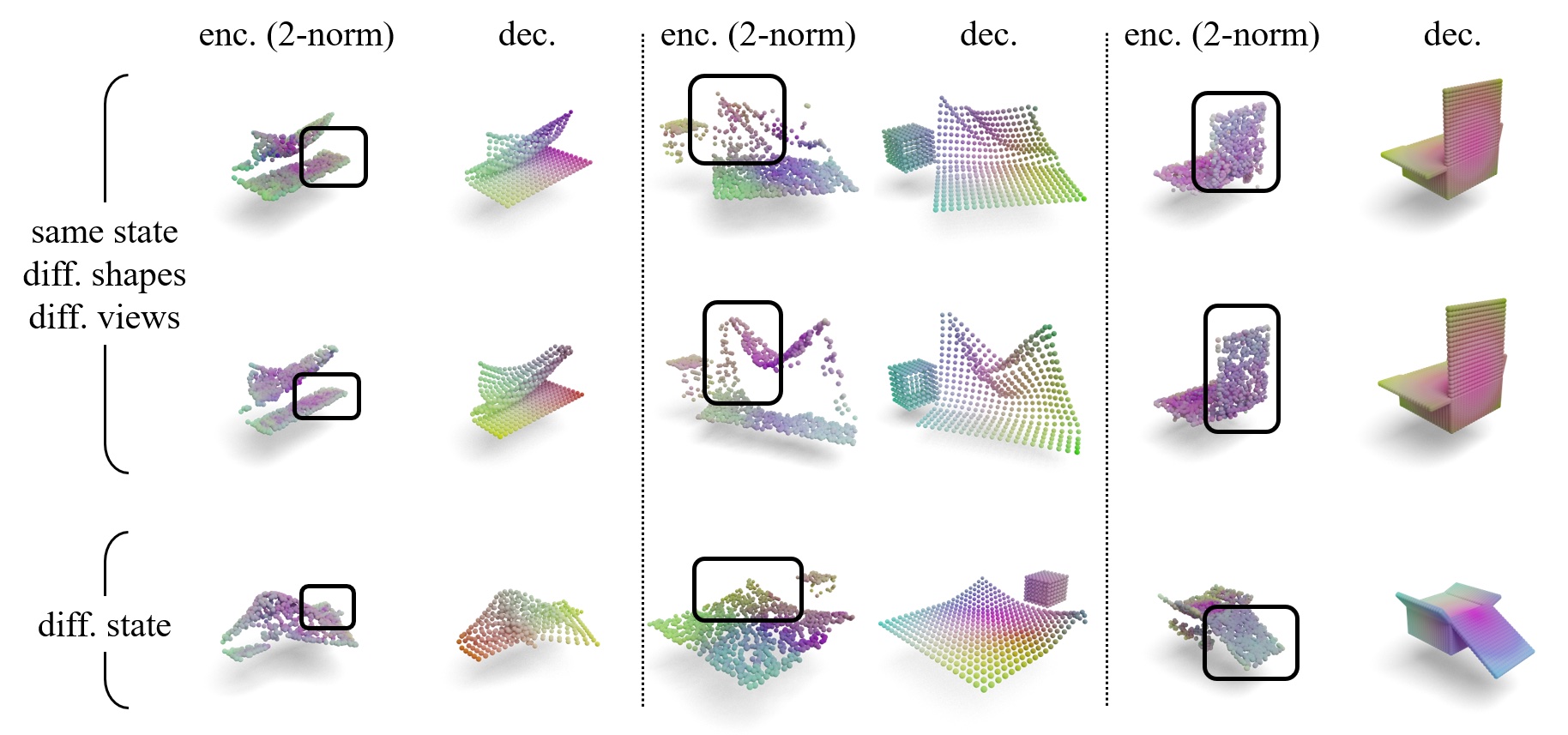}
    \caption{
    \textbf{Visualizations of per-point features.}
    The encoder features are equivariant vector-valued features on the partial point cloud observations and the visualizations are done on their invariant components (channel-wise 2-norms). The decoder features are invariant scalar-valued features on the complete objects. The RGB values are computed via PCA within each task. All point clouds are aligned to the canonical pose for visualization.
    \textbf{Top two rows:} Objects of different shapes viewed from different camera angles but at the same poses. Both encoder and decoder features show strong correspondences within each state due to the contrastive learning. \\
    \textbf{Bottom row:} Objects in a different state (features become different from the two top rows).
    }
    \label{fig:feat_vis}
    \vspace{-12pt}
\end{figure}

The first phase of our method is a representation learning phase that addresses two issues. First, although architecture based on vector neurons is equivariant to uniform scaling, we need to also handle non-uniform re-scaling of objects. Second, although rotation is handled via equivariance in principle, in practice we need robustness to nonlinear changes caused by objects' self-occlusions when altering camera view angles. Hence, we start with the representation learning phase, illustrated in Fig.~\ref{fig:rep_learning}, which helps ensure generalization to varied objects in the same category (e.g. non-uniform re-scaling) and ensures robustness to nonlinear effects of changing viewpoints.
We assume access to a dataset $\mathcal{D}_{\text{sim}} = \{ (\rmX^q \in \mathbb{R}^{N \times 3}, \rmM^q \in \mathbb{R}^{M \times 3}) \}_{q=1}^{|\mathcal{D}_{\text{sim}}|}$ containing point clouds of simulated scenes that include the objects of interest. In each sample, we record the partial point cloud of the scene unprojected from one camera view $\rmX^q$ and the ground-truth mesh of the objects in the scene $\rmM^q$. Since simulated scenes include full information about object meshes, we can assume that all ground-truth meshes $\rmM^q$ have the same number of vertices, and the points in two meshes correspond to each other. This canonicalization simplifies constructing effective losses for this learning phase.

We train the visual representation using a contrastive loss similar to \cite{xie2020pointcontrast}. Specifically, we sample a simulated data pair of object point clouds $(\rmX^1, \rmM^1)$ and $(\rmX^2, \rmM^2)$ where two different object instances share a similar pose, articulation, or deformation. Then, we aim to learn per-point features $\{z(\rvm^1_i)\}_{i=1}^{M}$ and $\{z(\rvm^2_i)\}_{i=1}^{M}$ at ground-truth mesh points. Since we know the two ground-truth meshes have one-to-one correspondences, we know that $z(\rvm^1_i)$ and $z(\rvm^2_i)$ correspond to the same points on the mesh and should have similar latent features. We then apply a PointInfoNCE loss \cite{xie2020pointcontrast} to train the latent representation:
\begin{equation}
    \gL = -\sum_{i=1}^M \log \frac{\exp(z(\rvm^1_i) \cdot z(\rvm^2_i)/\tau)}{\sum_{j \neq i} \exp(z(\rvm^1_i) \cdot z(\rvm^2_j)/\tau)}.
\end{equation}

Instead of the SDF decoder in \cite{lei2023efem} which is incompatible with thin-shell deformable objects with highly varying geometries (like cloth), or the contrastive loss on the multi-views of the same object as originally introduced in \cite{xie2020pointcontrast}, we sample positive point pairs across different objects at comparable poses. This encourages representation to learn similar features for corresponding points of varied object instances (see Fig. \ref{fig:feat_vis}).

\subsection{Learning Generalizable Visuomotor Policies with SIM(3)-equivariant Architecture}

\begin{figure}[t]
    \vspace{5px}
    \centering
    \includegraphics[width=0.485\textwidth]{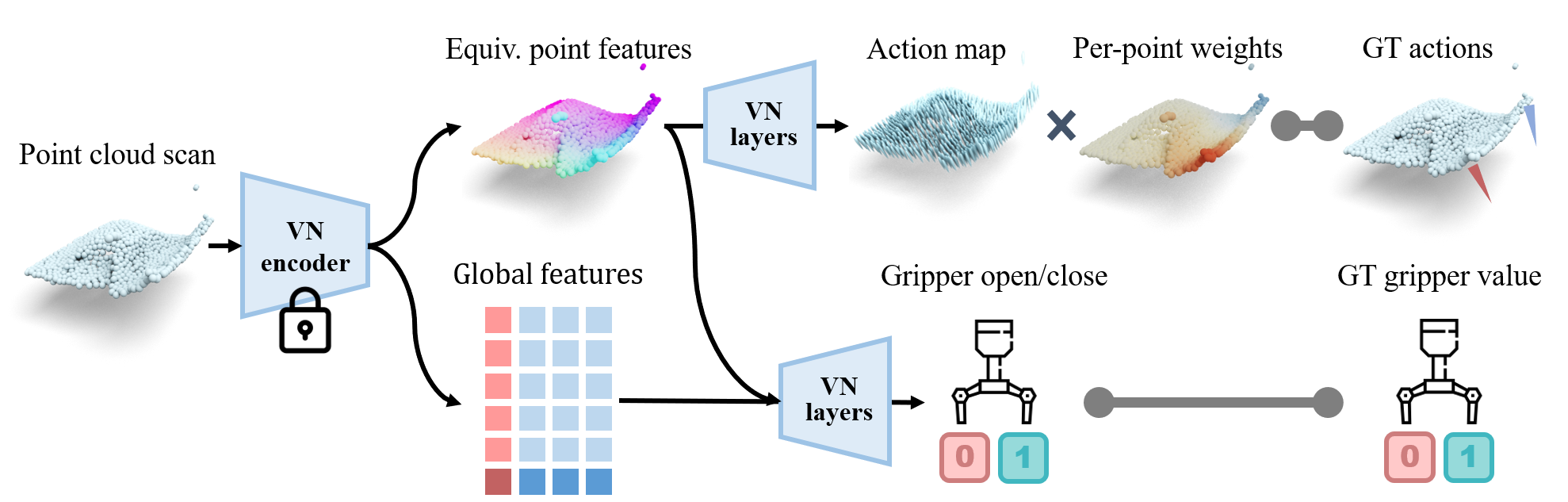}
    \caption{\textbf{Policy learning architecture.}
    We first pass a point cloud captured during policy execution through the frozen encoder from the representation learning phase to get local and global equivariant features. These are passed to two VN heads to get target end-effector velocities and open/close actions.}
    \vspace{-10pt}
    \label{fig:inference}
\end{figure}

After the representation learning phase, we freeze the encoder $\Phi$ and learn an action prediction head on top of the learned feature representation with a small number of human demonstrations.
As observed in~\cite{chen2020simple}, the intermediate representations from the encoder can possess desirable properties even without decoding at inference time.

\textbf{Architecture.} 
The policy learning architecture is illustrated in Fig. \ref{fig:inference}. Given input observation $o = (\rmX, \{\rvy^e\}_{e=1}^E)$, our policy $\pi$ first passes the input point cloud through the encoder $\Phi$ to obtain a global feature $\Theta_R$ and per-point local features $\{\rmV(\rvx_n)\}_{n=1}^N$. Then, the policy architecture is split into two branches. The end-effector velocity prediction branch takes the local features as input and passes them through three VN layers to obtain a velocity map $\{\hat{\rvv}_n\}_{n=1}^N$. When the policy is deployed, for each end-effector $e \!\in\! \{1, \ldots, E\}$ we find a point $\rvx_{k^e}$ in the point cloud that is closest to the current end-effector position $\rvy^e$; we use the velocity map output $\hat{\rvv}_{k^e}$ as the target velocity. 
The other branch passes the global and local features through three other VN layers to produce a target open/close command $\{\hat{g}^e\}_{e=1}^E$ for each end-effector.
The target end-effector velocities and gripper open/close commands are then concatenated to form the output action $\hat{a} = (\{\hat{\rvv}_{k^e}\}_{e=1}^E, \{\hat{g}^e\}_{e=1}^E)$.

\textbf{Loss function.}
We use a multi-component loss function for training the target velocities and gripper open/close commands. 
To supervise the outputs of the velocity map, we want to apply a loss term on every point in the point cloud. However, points that are located farther away from any end-effector in the demonstration should receive less supervision than points closer to one of the end-effectors. To achieve this, we weigh the loss term applied to each point by how far that point is from the end-effectors.
Concretely, the loss is defined by: 
\begin{equation}\mathcal{L}_\text{vel} = \frac{1}{N} \sum_{n=1}^N \sum_{e=1}^E  w(\rvx_n, \rvy^e) \cdot |\!|\hat{\rvv}_n - \rvv^e|\!|_2^2
\end{equation}
Here, $\rvv^e$ denotes {the ground truth velocity of the $e$-th end-effector, $\hat{\rvv}_n$ denotes the $n$-th point in the output velocity map}, and $w(\rvx_n, \rvy^e) = \exp(-(\rvx_n- \rvy^e)^2 / (2\sigma))$ is a weighting function that is large when $\rvx_n$ is close to $\rvy^e$ and small when $\rvx_n$ is far from $\rvy^e$.
The gripper open/close loss is simply defined as the mean squared error (MSE) between the gripper command $\hat{g}^e$ and the ground truth gripper state $g^e$ observed during the demo: $\mathcal{L}_{\text{grip}} = \text{MSE}(\hat{g}^e, g^e)$.
The policy learning loss is a weighted sum of the above loss components: \begin{equation} \mathcal{L} = \mathcal{L}_\text{vel} +  \lambda_\text{grip}\mathcal{L}_\text{grip}, \end{equation} where $\lambda_\text{grip}$ is a weighting term for the velocity component of the loss function, which we set to $0.05$ in our experiments to scale the two terms to comparable magnitudes.

\section{Experiments}

Through our experiments, we aim to answer the following questions: \textbf{(Q1)} Can our method learn from a few demonstrations in one scenario, then generalize to scenarios with unseen object sizes and poses? \textbf{(Q2)} Does our method outperform methods that rely on extensive augmentations to achieve generalization? \textbf{(Q3)} Does our method perform better than methods that don't employ equivariant architectures or representations pre-trained from simulation?

\subsection{Comparisons}

We compare our method with several baselines and also analyze ablations, as described below.

(1) \textbf{PointNet+BC:} a baseline behavior cloning (BC) algorithm that trains a neural network architecture, which takes a partial point cloud of the scene as input, encodes the input with a PointNet \cite{qi2017pointnet}, concatenates the point cloud feature with robot proprioception, then passes the concatenated vector through an MLP to obtain target robot actions. Since this baseline has no representation learning phase, it only uses the source task demos and does not utilize simulated data. This baseline examines a basic policy learning method one would use for processing point cloud inputs and outputting target actions without adding more complexity to handle generalization.

(2) \textbf{PointNet+BC$_{\text{aug}}$:} a variation of \textit{PointNet+BC} that augments the demonstration data with additional object poses and scales, aiming to make test-time scenarios `in distribution' with respect to the augmented data. The total amount of data used after augmentation equals the amount of data used in our method. This baseline serves to answer (Q2) from above.

(3) \textbf{PointContrast+BC:} this baseline first trains a latent representation of the 3D scene with PointContrast \cite{xie2020pointcontrast}, then trains an MLP that takes the learned features and robot proprioception as input and outputs target robot actions. This baseline serves to answer question (Q3).

(4) \textbf{PointContrast$_{\text{aug}}$+BC:} a variation of \textit{PointContrast+BC} that uses augmented object poses during feature training, which is a way of approximating equivariant features with Siamese training \cite{sun2021canonical}.

(5) \textbf{Ours w/o pre-training:} this ablation trains the whole architecture, including the visual encoder and the action heads, together without pre-training.

(6) \textbf{Ours w/o equivariance:} this ablation replaces all SIM(3)-equivariant parts in our method with non-equivariant PointNet-based components with similar sizes.

(7) \textbf{Ours w/o equivariance with aug:} this ablation is our method without  SIM(3)-equivariance, but with using augmentation for object poses and scales during feature training.

\subsection{Simulation Experiments}

\begin{figure}[t]
\vspace{5px}
    \centering
    \includegraphics[width=0.45\textwidth]{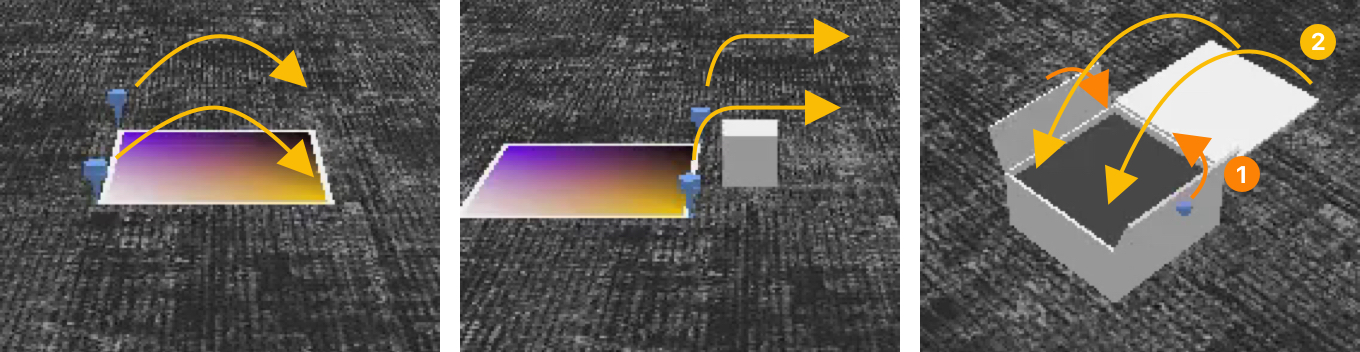}
    \caption{\textbf{Simulation environments.} \textit{Cloth Folding (left):} two grippers fold a piece of cloth by grasping two corners of the cloth. \textit{Object Covering (middle):} two grippers pick up a cloth at two corners and drag it to fully cover another object. This task tests the ability to handle scenarios with several objects. \textit{Box Closing (right):} two manipulators close a box with three flaps by first closing the side flaps and then closing the larger front/back flap. This task tests manipulation with articulated objects.}
    \label{fig:sim_env}
    \vspace{-15px}
\end{figure}

\textbf{Tasks.} We evaluate our method on three challenging robot manipulation tasks involving various deformable and articulated objects: cloth folding, object covering, and box covering. We illustrate the three environments in Fig. \ref{fig:sim_env}.

\begin{figure*}[t]
    \vspace{5px}
    \centering
    \includegraphics[width=0.9\textwidth]{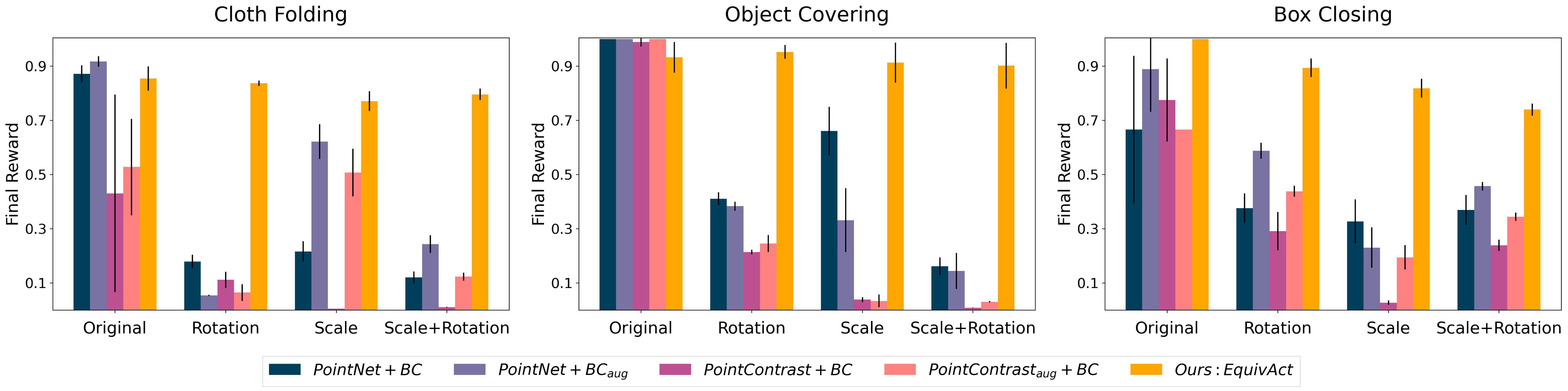}
    \caption{
    \textbf{Results for simulation experiments.}
    We evaluate 3 manipulation tasks involving deformable and articulated objects. The comparisons with baselines show that our method outperforms prior methods that rely on augmentations to achieve generalization or utilize non-equivariant representations.}
    \label{fig:sim_results}
    \vspace{-10px}
\end{figure*}

\begin{figure*}[t]
    \vspace{5px}
    \centering
    \includegraphics[width=0.9\textwidth]{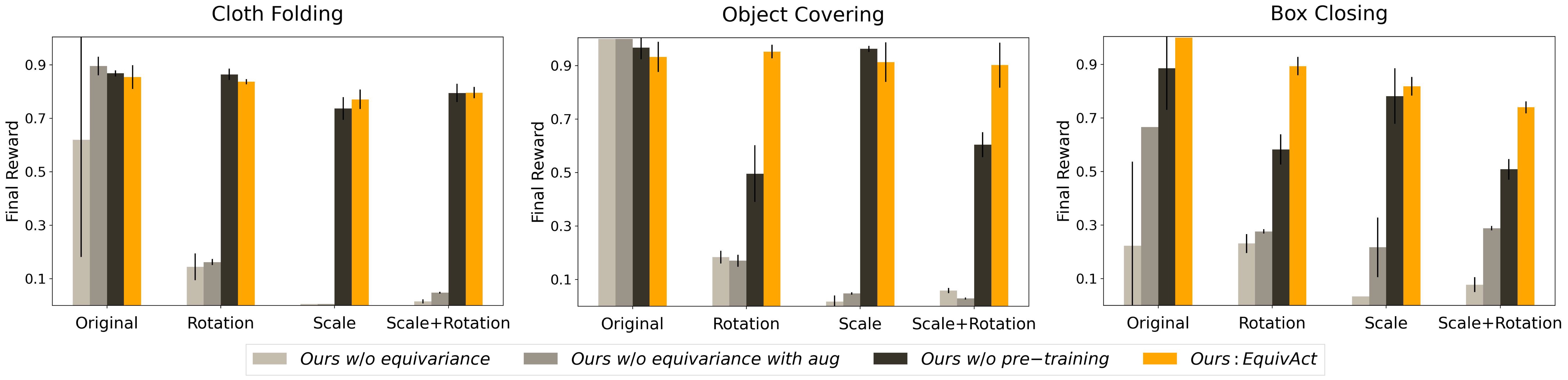}
    \caption{
    \textbf{Ablation results in simulation.}
    Comparisons show that the pre-training and equivariance components are essential for good performance.}
    \label{fig:abl}
    \vspace{-10px}
\end{figure*}

\textbf{Representation learning data.} 
Our simulation environments use PyBullet simulation engine~\cite{coumans2019}. For simplicity, we use floating end-effectors, which can physically push objects and attach to objects to imitate grasping. 
We generate randomized simulation data in each task for representation pre-training. In \textit{Cloth Folding} and \textit{Object Covering} tasks, we generate representation learning data by grasping two corners of the object and performing a pick-and-place motion to random target positions. We collected 200 simulated episodes for each task, varying the non-uniform scaling and camera viewpoint in each episode. Note that these episodes were not demonstrations and didn't have to include successful executions of the target task. In the \textit{Box Closing} task, we generated data for pre-training by taking snapshots of the box in various articulated poses (i.e. randomized positions for each box flap). We collected 20k point clouds (equivalent to 200 episodes, since episodes in \textit{Box Closing} are roughly 100 steps) with varying non-uniform scaling and camera pose.

\textbf{Demonstration data.} We collect 50 noisy demonstrations for all tasks. In each task, the 50 demos include the same object at the canonical pose. This means the policy is trained using demos of only one given manipulation scenario. 

\textbf{Evaluation.} To thoroughly examine the generalization capability of the learned policy, we evaluate it in four different setups. The \textit{Original} setup evaluates the performance of the policy with an object that is placed and sized the same as in the demonstrated scenario. The \textit{Rot} setup randomizes the object rotation while keeping the object scale constant. The \textit{Scale} setup tests non-uniform re-scaling, where each object dimension is re-scaled up to twice its original size, while keeping the aspect ratio within at most $1\!:\!1.33$ for the shortest vs. longest dimension; in this setup we keep the object in the canonical pose. The \textit{Rot+Scale} setup randomizes both the rotation and non-uniform scaling of the object. Note that except for the \textit{Original} setup, all other setups test out-of-distribution performance of the trained policy. In particular, in the \textit{Box Closing} task, we add a further challenge to the trained policy by removing the contents in the box at test time, which introduces further differences in point clouds between train and test time for this task.
We measure the performance of a task using a task-specific reward function scaled between 0 and 1. In the \textit{Cloth Folding} task, the reward is measured by how close the bottom two corners of the cloth are to the top two corners. In the \textit{Object Covering} task, the reward equals the fraction of the object volume within the convex hull of the final cloth mesh. In the \textit{Box Closing} task, the reward equals $\frac{\theta_{\text{left}} + \theta_{\text{right}} + \theta_{\text{top}}}{3 \pi}$, where $\theta_{\text{left}}$, $\theta_{\text{right}}$, and $\theta_{\text{top}}$ denote the angular pose of the left, right, and top flaps (0 denotes fully open and $\pi$ denotes fully closed).

\textbf{Results.} Fig. \ref{fig:sim_results} shows experimental results in simulation. All methods perform well in the \textit{Original} evaluation setup. This shows that all methods learn to complete the demonstrated tasks. In the other setups, the two \textbf{PointNet+BC} baselines display a significant performance drop. Even after adding augmentations in \textbf{PointNet+BC$_{\text{aug}}$}, the baseline is still not able to recover to the level of its `in-distribution' performance for any of the three tasks. This shows that augmentation does not guarantee effective generalization to varied object scales and poses. Similarly, the \textbf{PointContrast+BC} baseline performs well in the \textit{Rot} setup but struggles to perform well in the evaluation setups that involve scaling. In contrast,  our pre-training phase that utilizes paired data with non-uniform re-scaling helps our method perform well across all setups, including \textit{Scale} and \textit{Rot+Scale} (with only a minor performance drop in the simpler \textit{Original} setup overall).

Our ablation experiments are presented in Fig. \ref{fig:abl}. The drop in performance of \textbf{Ours w/o Pretraining} and \textbf{Ours w/o Equivariance} ablations show that pre-training and equivariance are essential to prevent performance degradation.
In particular, \textbf{Ours w/o Pretraining} ablation performs substantially worse than our method in the \textit{Rotate} setup in the covering and closing tasks. In these tasks, different viewing angles can cause significant differences in point cloud observations, e.g. different self-occlusions of the object (though this effect is not significant in the folding task). This ablation result shows that the pre-training phase effectively makes our method robust to nonlinear changes in point cloud observations caused by varied viewing angles. 

\subsection{Real Robot Experiments}

To illustrate the effectiveness of our method in the real world, we test it with both quantitative and qualitative experiments in a mobile manipulation setup. 

\begin{figure}[t]
    \vspace{5px}
    \centering
    \includegraphics[width=0.37\textwidth]{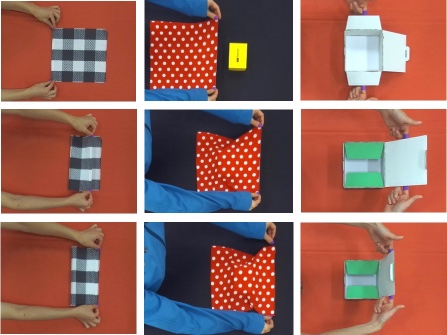}
    \caption{\textbf{Human demonstrations collected for real robot experiments.} From left to right: cloth folding, object covering, and box closing tasks.}
    \vspace{-10px}
\end{figure}

\begin{table}[b]
\vspace{-5px}
\centering
\begin{tabular}{@{}lccc@{}}
\toprule
                & Cloth Folding & Object Covering & Box Closing \\ \midrule
PointNet+BC$_{\text{aug}}$ & 0.083         & 0.000           & 0.000       \\
Ours (EquivAct) & \textbf{0.919}         & \textbf{0.825}           & \textbf{0.867}     \\\bottomrule
\end{tabular}
\caption{\textbf{Quantitative results for real robot experiments.} The table reports average performance over 10 trials. We randomize the position of the objects in a $\pm 20cm$ range and the orientation of the objects in a $\pm 15^\circ$ range. In all tasks, our method can complete the target manipulation task in the majority of trials, while the baseline fails to finish all tasks. Fig.~\ref{fig:hwres} shows qualitative samples produced by our policy.}
\label{tab:real_robot_results}
\end{table}

\textbf{Tasks and data collection.} 
We run our method in the same three manipulation tasks as our simulation experiments. In each task, our method takes 20 human demonstrations collected on a tabletop setup as training data. 
We use a ZED 2 stereo camera positioned above the table to record the movement of the objects and human hands. After data collection, we segment out the part of the point cloud that corresponds to the objects the human is manipulating and also parse out the human finger positions in each frame. We treat segmentation as a separate research problem, which could be addressed by state-of-the-art methods, such as~\cite{kirillov2023segment}. To avoid conflating the evaluation of the performance of various segmentation methods in our experiments, we segment out the objects using simple (robust) color filtering techniques.

\textbf{Mobile robot setup.} 
The robots operate in a large $4 \times 3$ meter workspace. 
We use holonomic mobile bases with a powered-caster drive system~\cite{holmberg2000development}, and Kinova Gen3 7-DoF arms equipped with Robotiq 2F-85 grippers. We use 
\mbox{ZED 2} stereo cameras positioned on the ceiling to obtain point clouds of the scene, then segment out the relevant parts of the scene. The policy takes the partial point cloud of the objects in the scene and the proprioception readings of the two mobile robots as input, then outputs a velocity and gripper open/close command for each robot. 

\begin{figure}[t]
    \vspace{5px}
    \centering
    \includegraphics[width=0.37\textwidth]{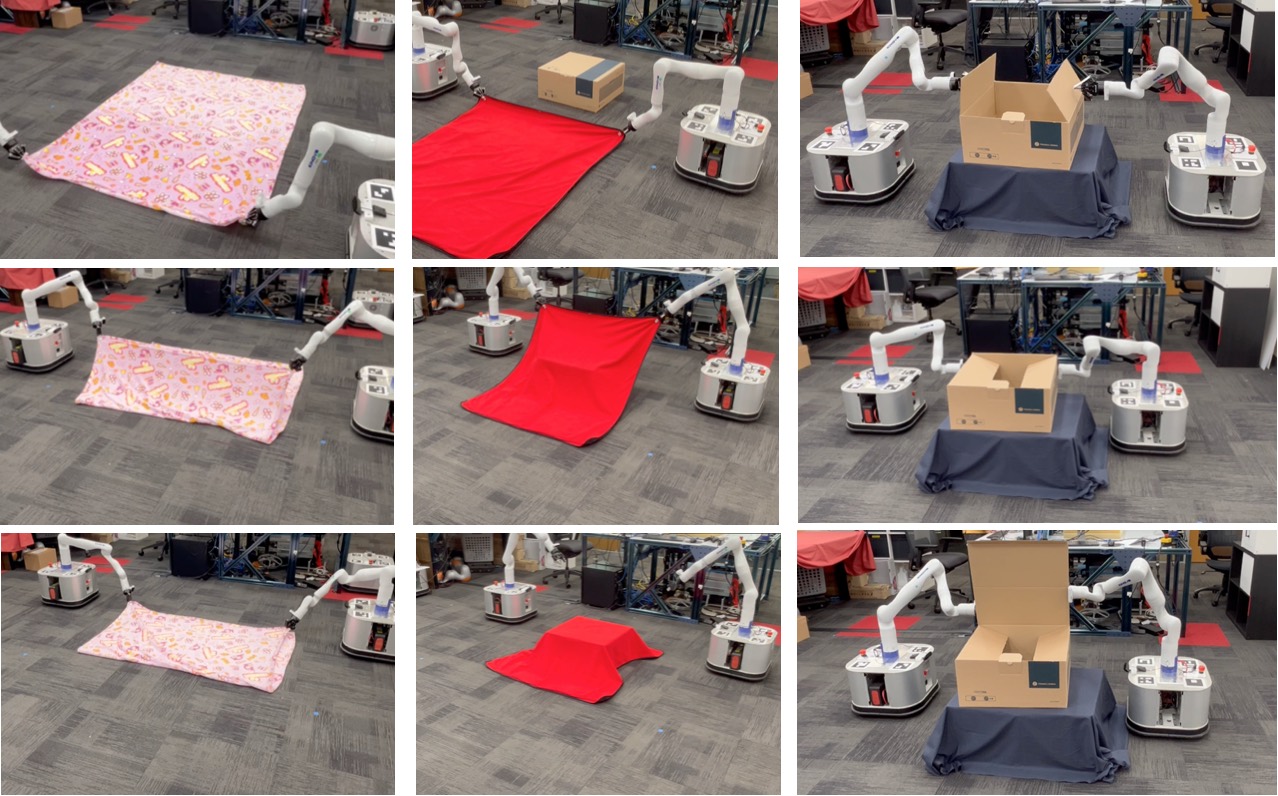}
    \caption{\textbf{Qualitative samples of deploying the policy on hardware.} We deploy our learned policies to a mobile robot platform to manipulate target objects that are at least 6 times larger than the demonstrated objects. They have different visual appearances, aspect ratios, and physical properties (e.g. friction and elasticity of the cloth) than those in the source (tabletop) tasks. Our policy can generalize zero-shot to the target objects and successfully carry out the task. Please see our \href{https://equivact.github.io}{website} for full videos of these samples.}
    \label{fig:hwres}
    \vspace{-5px}
\end{figure}

\textbf{Results.}
We compare our method with \textbf{PointNet+BC$_{\text{aug}}$}, the baseline that performed best in simulation. Table~\ref{tab:real_robot_results} shows quantitative results. 
We use similar evaluation metrics as those in the simulation experiments, with a few simplifications. In \textit{Object Covering}, we measure the fraction of the box covered as seen from the top. In \textit{Box Closing}, we count the number of box flaps fully closed (divided by the number of flaps) rather than computing exact angles.
In all three tasks, our method extrapolates from tabletop-scale demonstrations and successfully manipulates objects that are up to six times larger in size. In contrast, the \textbf{PointNet+BC$_{\text{aug}}$} baseline has a lot of trouble when attempting all three tasks. The common failure mode of the baseline is that it keeps producing target robot velocities that point in one direction. This is likely because the method is not able to generalize to out-of-distribution visual inputs.

\begin{figure}[t]
    \centering
    \includegraphics[width=0.37\textwidth]{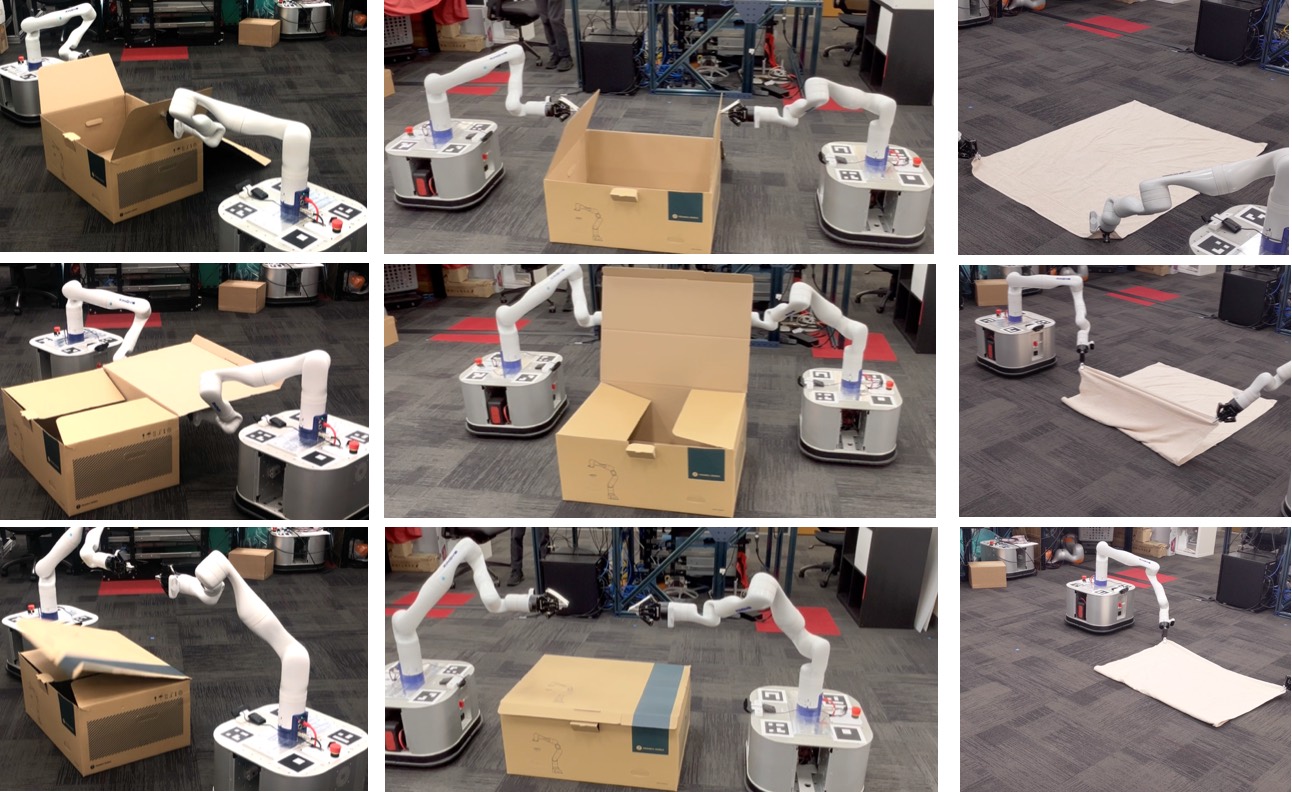}
    \caption{\textbf{Task variations.} To illustrate the generalization capability of our method, we test it on various objects and initial poses. The first two columns show that the \textit{Box Closing} policy can close different-sized boxes in different initial rotations. The third column shows that the \textit{Cloth Folding} policy can handle cloths with distinct appearance and scale compared to the pink blanket in Fig. \ref{fig:hwres}. See our \href{https://equivact.github.io}{website} for further examples.}
    \vspace{-15px}
    \label{fig:hwvar}
\end{figure}

We further demonstrate the robustness of our method by varying the position, rotation, scale, and appearance of the objects (see Fig. \ref{fig:hwvar}). 
Our method can robustly generalize to all these variations.
\section{Conclusion}

We presented \name, a visuomotor policy learning method that learns generalizable closed-loop policies for 3D manipulation tasks. 
We showed that our method successfully generalizes from a small set of demonstrations to a diverse set of target scenarios in both simulation and real robot experiments.
We hope that our work motivates further use of equivariant architectures and facilitates learning generalizable robot manipulation policies.

\newpage
\balance
\bibliographystyle{references/IEEEtran}
\bibliography{references/references}

\end{document}